\documentclass{article}
\usepackage{spconf,amsmath,graphicx}
\usepackage{multirow}
\usepackage{enumitem}
\usepackage{booktabs}
\setlist{nosep, leftmargin=14pt}
\usepackage{url}

\usepackage{mwe} 
\usepackage{graphicx}

\title{Simultaneous q-Space Sampling Optimization and Reconstruction for Fast and High-fidelity Diffusion Magnetic Resonance Imaging}

\name{Jing Yang$^{1,2}$, Jian Cheng$^{3,6}$, 
 Cheng Li$^{1}$, Wenxin Fan$^{1,2}$, Juan Zou$^{1}$, Ruoyou Wu$^{1,2,5}$ and Shanshan Wang$^{1,2,4,5}$ }

\address{$^{1}$Paul C. Lauterbur Research Center for Biomedical Imaging, Shenzhen \\Institute of Advanced Technology, Chinese Academy of Sciences, Shenzhen, Guangdong, China \\
    $^{2}$University of Chinese Academy of Sciences, Beijing, China\\
    $^{3}$State Key Laboratory of Software Development Environment, Beihang University, Beijing China. \\
    $^{4}$Peng Cheng Laboratory, Shenzhen, Guangdong, China\\
    $^{5}$Guangdong Provincial Key Laboratory of Artificial Intelligence in Medical Image,\\ Analysis and Application, Guangdong, China\\
    $^{6}$Key Laboratory of  Data Science and Intelligent Computing,\\Institute of International Innovation, Beihang University, Yuhang District, Hangzhou, China
}
 
\begin{document}
%
\maketitle
\begin{abstract}
Diffusion Magnetic Resonance Imaging (dMRI) plays a crucial role in the noninvasive investigation of tissue microstructural properties and structural connectivity in the \textit{in vivo} human brain. However, to effectively capture the intricate characteristics of water diffusion at various directions and scales, it is important to employ comprehensive q-space sampling. Unfortunately, this requirement leads to long scan times, limiting the clinical applicability of dMRI. To address this challenge, we propose SSOR, a Simultaneous q-Space sampling Optimization and Reconstruction framework. 
We jointly optimize a subset of q-space samples using a continuous representation of spherical harmonic functions and a reconstruction network. Additionally, we integrate the unique properties of diffusion magnetic resonance imaging (dMRI) in both the q-space and image domains by applying $l1$-norm and total-variation regularization.The experiments conducted on HCP data demonstrate that SSOR has promising strengths both quantitatively and qualitatively and exhibits robustness to noise.

\end{abstract}
\section{Introduction}
\label{sec:intro}
Diffusion Magnetic Resonance Imaging (dMRI) is an important technology for noninvasively characterizing the microstructure of the brain tissue, particularly white matter \cite{zheng2023microstructure}. By capturing and modeling the Brownian motion of water molecules, dMRI provides valuable insights into tissue properties
\cite{mori1999diffusion, taouli2010diffusion}. The versatility and broad applicability of dMRI have made it an indispensable tool in both clinical applications and neuroscience research, spanning across diverse areas such as stroke prediction, tumor detection, neuroscience studies, and myocardial microstructure imaging \cite{marini2007quantitative, tong1998correlation, wu2007mr}.
However, to effectively capture the intricate patterns of water diffusion across diverse directions and scales, as well as to ensure precise tissue microstructure model fitting and subsequent quantitative analysis, dense sampling in the diffusion gradient space, commonly referred to as q-space, is necessary for dMRI \cite{chen2018angular}. Unfortunately, the acquisition time of dMRI significantly increases with the increasing number of sampled diffusion-encoding directions, which limits its applicability in clinical practice. Therefore, it is highly desirable to develop fast dMRI approaches that can enable high angular resolution imaging from sparse q-space samples. 

The fast development of deep learning techniques opens up a promising avenue for fast dMRI. In recent years, several q-space deep learning methods have been developed. For example, Golkov et al. \cite{golkov2016q} proposed to directly fit dMRI signals to diffusion parameter maps using a simple three-layer multilayer perceptron (MLP). Yin et al. \cite{yin2019fast} introduced a one-dimensional encoder-decoder convolutional network for rapid reconstruction of high angular resolution diffusion imaging (HARDI) signals. 
Nevertheless, in deep learning-based fast imaging methods for structural MRI and dMRI, most of the sampling trajectories are predetermined, using either fixed or random masks \cite{wang2021deep}. There are only a few studies performing joint k-space sampling optimization and image reconstruction to simultaneously learn the most suitable sampling trajectory and improve the quality of reconstructed images 
Sanchez et al.  \cite{sanchez2020scalable} solved the problem of sampling mask optimization for arbitrary reconstruction methods and limited acquisition time by finding an optimal probability distribution and drawing a mask with fixed cardinality. 
Weiss et al. \cite{weiss2021towards} proposed a method that focuses on optimal q-space sampling in dMRI. 
While existing deep learning methods for rapid dMRI have shown promising results, they often suffer from a limitation in adequately incorporating q-space information and spatial domain information into their models and loss function designs \cite{golkov2016q, jha2020multi}. To this end, we propose an innovative approach called SSOR (simultaneous q-space sampling optimization and reconstruction framework) to accelerate dMRI. The overall framework of SSOR is illustrated in Figure 1. Specifically, SSOR employs deep learning techniques to jointly optimize the sampling points in the q-space and reconstruct the high angular resolution dMRI signal, which fully utilizes the information in both the q-space and image domain. 
The main contributions of this work can be summarized as follows:
\begin{enumerate}
    \item We develop a simultaneous q-space sampling optimization and reconstruction framework with a specifically designed loss function, exploiting the information in the q-space domain and the image domain for accelerated dMRI.
    \item The proposed framework enables highly accurate reconstruction of dMRI data acquired from various q-space sampling protocols, surpassing the performance of existing competitive methods.  
    \item The framework is flexible and adaptable to different q-space sampling protocols, allowing it to be tested with sampling protocols different from those utilized for model training. This feature enhances the applicability of the model in prospective imaging acquisition scenarios.   
\end{enumerate}




\section{Method}
\label{sec:method}

\subsection{q-Space Sampling Optimization block}
\label{ssec:subhead1}
The dMRI data are 4D signals $x \in R^{m\times n\times s\times d}$, where $m$, $n$, $s$, and $d$ refer to the weight, height, slice, and diffusion gradient direction, respectively. Each diffusion data can be considered as a set of 3-dimension volumes ($m\times n\times s$) captured in the q-space. 
The sparse sampling process is accomplished using spherical harmonics, which allows for a uniform and standardized representation of the DWI data and maximizes the utilization of information in q-space. According to the previous work \cite{descoteaux2007regularized}, for DWI signal at each of the N gradient directions $i$, we have:
\begin{equation}
S(\theta_i, \phi_i)=\sum_{j=1}^R c_j Y_j(\theta_i, \phi_i),
\end{equation}
\begin{figure}[htp]
  \centering
  \includegraphics[width=0.5\textwidth]{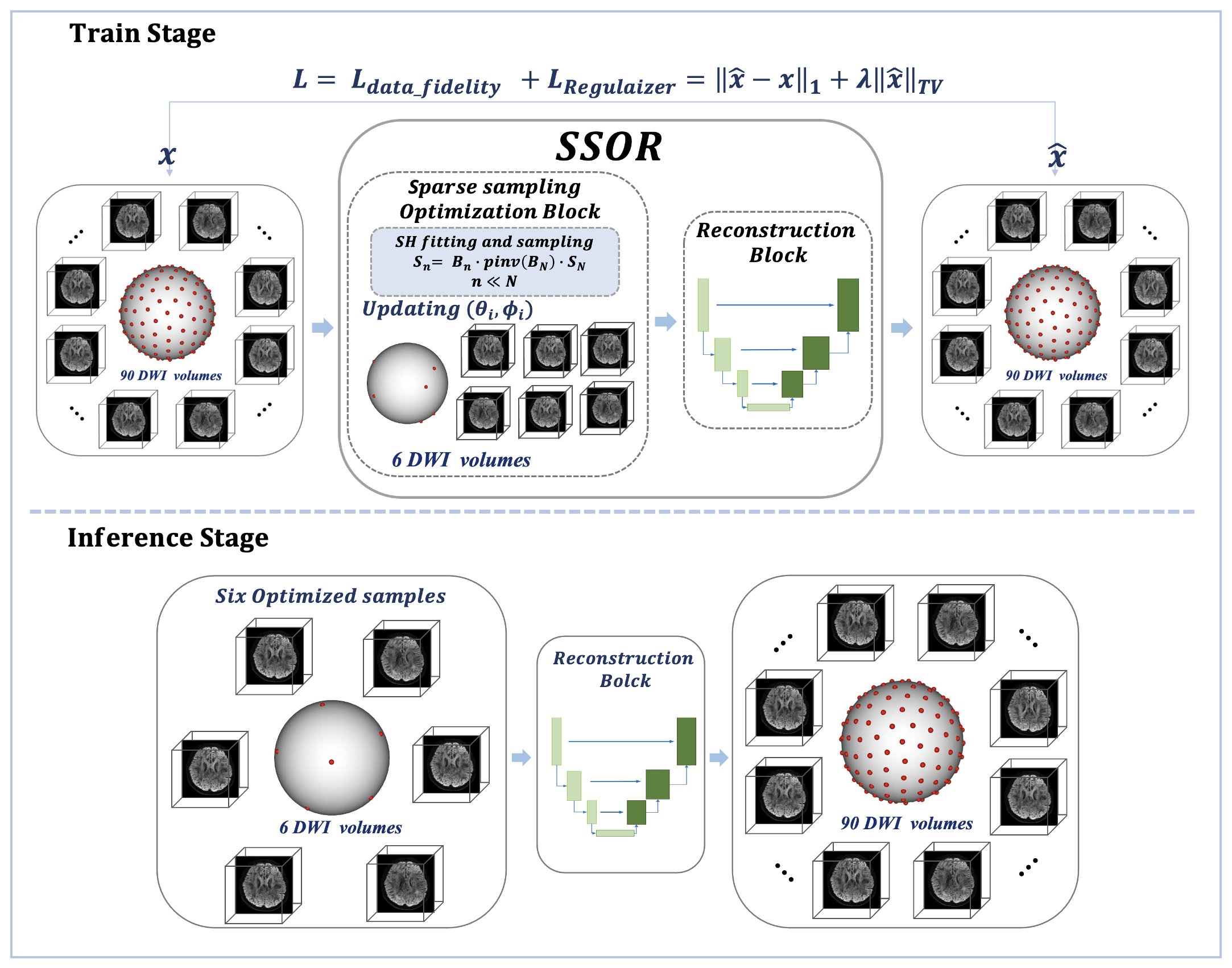}
  \caption{The illustration of SSOR framework.}
\end{figure}
where $(\theta_i, \phi_i)$ are standard polar angles ($\theta_i \in(0, \pi)$, $\phi_i \in(0,2 \pi)$). $Y_j$ is the $j^{th}$ Spherical Harmonics (SH) basis and $c_j$ is the corresponding SH coefficient and $\mathrm{B}_{\mathrm{N} \times \mathrm{R}}$ is the number of terms in the real SH basis $Y$ of order $l$. Setting $S_{N}$ to be a $\mathrm{N} \times{1}$ vector (representing the N encoding gradient direction), $C_{N}$ is the $\mathrm{R} \times{1}$ vector of SH coefficients $c_j$ and $B_{N}$ is the $\mathrm{N} \times \mathrm{R}$ matrix constructed with the discrete real SH basis, which can be defined as: 
\begin{equation}
B_{N}=\left[\begin{array}{ccc}
Y_1\left(\theta_1, \phi_1\right) & \cdots & Y_R\left(\theta_1, \phi_1\right) \\
\vdots & \ddots & \vdots \\
Y_1\left(\theta_N, \phi_N\right) & \cdots & Y_R\left(\theta_N, \phi_N\right)
\end{array}\right].
\end{equation}


We can get the linear system $\mathbf{S_{N}}=\mathbf{B_{N}}\mathbf{C_{N}}$, and the sparse sampled DWI signal can be define as:
\begin{equation}S_{n}=B_{n}\cdot C_{N},
\end{equation}
where 
\begin{equation}
C_{N}=pinv(B_{N})\cdot S_{N},
\end{equation}
and
\begin{equation}B_{n}=\begin{bmatrix}Y_1(\theta_1',\phi_1')&\cdots&Y_R(\theta_1',\phi_1')\\\vdots&\ddots&\vdots\\Y_1(\theta_n',\phi_n')&\cdots&Y_R(\theta_n',\phi_n')\end{bmatrix}.\end{equation}

Through the joint sparse sampling optimization and reconstruction framework, we can get learnable spherical coordinates $\mathrm{\varphi^{\prime}}$ and $\theta^{\prime}$ in $B_{n}$ and optimize the sparse sampling directions ($\mathrm{n\ll N}$) during framework training.


\subsection{Reconstruction block}
\label{ssec:subhead2}
These sparse-sampled data are then fed into the reconstruction network for joint sampling optimization and reconstruction network training, employing an elaborately designed loss function. Through network training and optimization, optimal q-space sparse sampling points $\mathrm{\varphi^{\prime}}$ and $\theta^{\prime}$ are obtained. Our framework is flexible in terms of the reconstruction model used. In this study, we adopt the U-net \cite{ronneberger2015u} architecture. 
\subsection{Training Loss}
\label{ssec:subhead3}
Inspired by the works of Ye et al. \cite{ye2017estimation,ye2020improved} which assume that the diffusion signal has a sparse representation in the q-space for the estimation of tissue microstructure and explore the sparsity of diffusion signals for q-DL, we employ end-to-end deep network based on q-space sparse representation to reconstruct high angular resolution dMRI data using $l_1$ norm. Total-variation regularization is also introduced to guarantee that more structure details can be restored\cite{ning2016joint}. The overall designed training loss is:
\begin{equation}
L=\left.\|\widehat{x}-x\|_1+\lambda\|\widehat{x}\|_{TV}\right.,
\end{equation}
where
\begin{equation}
\begin{gathered}
\hat{x}=R_\psi\left(Q_{\theta, \varphi}(x)\right)
\end{gathered},
\end{equation}

where $\hat{x}$ is the output of joint optimal sampling and reconstruction network $R_\psi\left(Q_{\theta, \varphi}(\cdot)\right)$ and $x$ is the high angular resolution dMRI.

\section{Experiments}
\label{sec:Exp}
We compared the performance of our method with learned-dMRI \cite{weiss2021towards}, reconstruction network with random sampling and reconstruction network with uniform sampling \cite{jones1999optimal} on data with different distributions.

\subsection{Datasets}
\label{ssec:Data}
Pre-processed diffusion MRI data from the publicly available Human Connectome Project (HCP) dataset were used for this study \cite{van2013wu}. The dMRI data were acquired in the whole brain at 1.25 mm isotropic resolution using a two-dimensional diffusion-weighted pulsed-gradient spin-echo echo-planar imaging (DW-PGSE-EPI) sequence, with four b-values (0, 1000, 2000, 3000 $\mathrm{s} / \mathrm{mm}^2$). For each non-zero b-value, 90 DWI volumes sampled along uniformly distributed diffusion-encoding directions were acquired \cite{caruyer2013design}. In total, data from 230 subjects were used, which were split into 175 subjects for training, 24 for validation, and 31 for testing.
\subsection{Implementation Details}
\label{ssec:details}
The q-space sampling stage and the reconstruction stage were trained with the Adam optimizer \cite{kingma2014adam}, using one Tesla V100 GPUs (32GB). The number of epochs for training was set to 50. The learning rate was empirically set to 1e-3 for the sampling stage and 1e-4 for the reconstruction stage. $\lambda$ was empirically set to 2e-7.
 The framework was implemented using PyTorch library. The reconstruction network was applied to each image slice and reconstruction results were concatenated to a 3D volume during the inference stage. We conducted experiments with the following acceleration factors (AF): 30, 15, and 10, which correspond to $n=$ 3, 6, and 9 q-space samples, respectively. 
\subsection{Results}
\label{ssec:results}
\subsubsection{Experimental results on in-distribution data }
\label{sssec:b1000 results}
\begin{figure}[htb]
  \centering
  \includegraphics[width=0.5\textwidth]{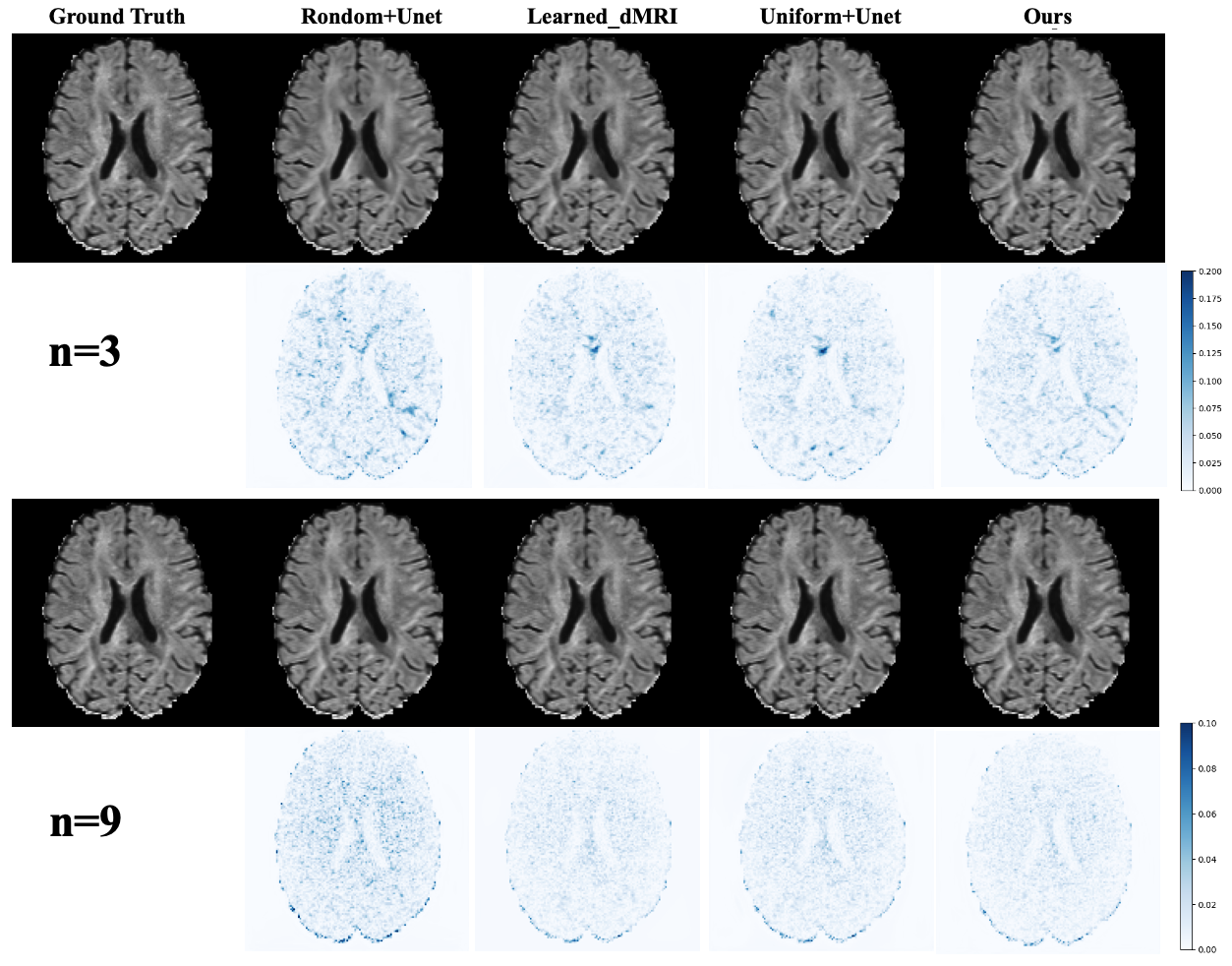}
  \caption{Qualitative results of different methods.}
  \label{fig2}
\end{figure}

Table \ref{tab1} shows the quantitative results on ($\mathbf{b}=1000\mathbf{s}/{\mathrm{mm}}^{2}$) data . The PSNR and SSIM of different methods show an increasing trend along with the increase of the sampled directions. The performance improvement of our method compared to the three comparison methods is obvious, especially in terms of SSIM. We randomly selected a slice from the ($\mathbf{b}=1000\mathbf{s}/{\mathrm{mm}}^{2}$)test set for qualitative evaluation (Figure \ref{fig2}). Visually, our results look more similar to the ground truth, demonstrating the enhanced reconstruction capability of our method.

\begin{table}[htp]
\begin{center}
\caption{Quantitative results of different methods on $\mathbf{b}=1000\mathbf{s}/{\mathrm{mm}}^{2}$ data.}
\label{tab1}
\resizebox{0.5\textwidth}{!}{
\begin{tabular}{c c c c c c c}
\hline
\multirow{2}{*}{Methods} & \multicolumn{3}{c}{PSNR} & \multicolumn{3}{c}{SSIM} \\  
\cline{2-7}
\multicolumn{1}{c}{} & n=3 & n=6 & n=9 & n=3 & n=6 & n=9 \\
\hline
Random+Unet   & 34.10 & 36.39 & 38.71 & 0.6518 & 0.7213 & 0.7831 \\
Uniform+Unet  & 35.58 & \textbf{39.05} & 38.88 & 0.7532 & 0.8504 & 0.7696 \\
Learned-dMRI  & 34.48 & 38.49 & 39.46 & 0.6442 & 0.7928 & 0.8028 \\
Ours &\textbf{36.19} & 38.91 & \textbf{39.70} &\textbf{0.8566} & \textbf{0.8918} & \textbf{0.8886} \\
\hline
\end{tabular}}
\end{center}
\end{table}

\begin{table}[htp]
\begin{center}
\caption{Quantitative results of different methods when tested on  $\mathbf{b}=2000\mathbf{s}/{\mathrm{mm}}^{2}$ data.}
\label{tab2}
\resizebox{0.5\textwidth}{!}{
\begin{tabular}{c c c c c c c}
\hline
\multirow{2}{*}{Methods} & \multicolumn{3}{c}{PSNR} & \multicolumn{3}{c}{SSIM} \\  
\cline{2-7}
\multicolumn{1}{c}{} & n=3 & n=6 & n=9 & n=3 & n=6 & n=9 \\
\hline
Random+Unet   & 33.25 & 35.61 & 37.92 & 0.7000 & 0.7737 & 0.8143 \\
Uniform+Unet  & 34.93 & \textbf{38.10} & 38.29 & 0.7953 & 0.8653 & 0.7893 \\
Learned-dMRI  & 33.80 & 37.61 & 38.73 & 0.7126 & 0.8281 & 0.8319 \\
Ours &\textbf{35.32} & 37.96 & \textbf{39.04} &\textbf{0.873} & \textbf{0.8867} & \textbf{0.9011} \\
\hline
\end{tabular}}
\end{center}
\end{table}
\begin{table}[ht]
\begin{center}
\caption{Quantitative results of different methods when tested on  $\mathbf{b}=3000\mathbf{s}/{\mathrm{mm}}^{2}$ data.}
\label{tab3}
\resizebox{0.5\textwidth}{!}{
\begin{tabular}{c c c c c c c}
\hline
\multirow{2}{*}{Methods} & \multicolumn{3}{c}{PSNR} & \multicolumn{3}{c}{SSIM} \\  
\cline{2-7}
\multicolumn{1}{c}{} & n=3 & n=6 & n=9 & n=3 & n=6 & n=9 \\
\hline
Random+Unet   & 32.20 & 34.52 & 36.62 & 0.6671 & 0.7526 & 0.7948 \\
Uniform+Unet  & 34.07 & \textbf{36.94} & 36.90 & 0.7687 & 0.8606 & 0.7668 \\
Learned-dMRI  & 32.90 & 36.42 & 37.47 & 0.6944 & 0.8177 & 0.8250 \\
Ours &\textbf{34.34} & 36.69 & \textbf{37.58} &\textbf{0.8605} & \textbf{0.8867} & \textbf{0.8925} \\
\hline
\end{tabular}}
\end{center}
\end{table}
\subsubsection{Experimental results on out-of-distribution data }
\label{sssec:b2000 and b3000 results}
Tables \ref{tab2} and \ref{tab3} show the reconstruction results using different acquisition protocols (i.e., different b values). Higher b values are more sensitive to the diffusion of water molecules, while signal-to-noise ratios of the data are lower. Here, we directly tested the models trained using the $\mathbf{b}=1000\mathbf{s}/{\mathrm{mm}}^{2}$ data on the $\mathbf{b}=2000\mathbf{s}/{\mathrm{mm}}^{2}$ and $\mathbf{b}=3000\mathbf{s}/{\mathrm{mm}}^{2}$ data. Overall, our model consistently performs better than the comparison models on the lower SNR ratio data, showing the robustness of our method to data noise.

\section{Conclusions}
\label{Con}
In this study, we propose SSOR, a simultaneous q-space optimization and reconstruction framework to accelerate dMRI. By leveraging the inherent sparsity properties of q-space and employing total-variation regularization in the image domain, our framework enables accelerated dMRI while maintaining robust reconstruction performance with highly sparse sampling in the q-space. It exhibits strong robustness when tested on lower SNR datasets. The strategy for selecting q-space samples in SSOR is both flexible and adaptable, allowing for direct application in prospective imaging acquisition. This approach holds great potential for improving the acquisition and reconstruction workflow of dMRI and advancing its clinical applicability.

\section{Acknowledgments}
\label{sec:acknowledgments}

This research was partly supported by the National Natural Science Foundation of China (62222118, U22A2040), Guangdong Provincial Key Laboratory of Artificial Intelligence in Medical Image Analysis and Application (2022B12\\12010011), Shenzhen Science and Technology Program (RCYX20210706092104034, JCYJ20220531100213029), and Key Laboratory for Magnetic Resonance and Multimodality Imaging of Guangdong Province (2023B1212060052).

\bibliographystyle{IEEEbib}
\bibliography{refs}

\end{document}